\documentclass[review]{elsarticle}

\usepackage{lineno,hyperref}
\modulolinenumbers[10]

\usepackage{geometry}
\usepackage[nodots]{numcompress}

\biboptions{numbers,sort&compress}

\usepackage{amsmath}
\usepackage[nodots]{numcompress}
\usepackage{textcomp}
\biboptions{numbers,sort&compress}

\journal{arXiv}

\begin{document}

\begin{frontmatter}

\title{Indoor simultaneous localization and mapping based on fringe projection profilometry}


\author[mymainaddress,mysecondaryaddress]{Yang Zhao}

\author[mymainaddress,mysecondaryaddress]{Kai Zhang}

\author[mymainaddress,mysecondaryaddress]{Haotian Yu}

\author[mymainaddress,mysecondaryaddress]{Yi Zhang}

\author[mymainaddress,mysecondaryaddress]{Dongliang Zheng\corref{mycorrespondingauthor}}
\cortext[mycorrespondingauthor]{Corresponding author}
\ead{dlzheng@njust.edu.cn}

\author[mymainaddress,mysecondaryaddress]{Jing Han\corref{mycorrespondingauthor}}
\ead{eohj@njust.edu.cn}

\address[mymainaddress]{School of Electronic and Optical Engineering, Nanjing University of Science and Technology, No. 200 Xiaolingwei Street, Nanjing, Jiangsu Province 210094, China}
\address[mysecondaryaddress]{Jiangsu Key Laboratory of Spectral Imaging and Intelligent Sense, Nanjing University of Science and Technology, Nanjing, Jiangsu Province 210094, China}

\begin{abstract}
Simultaneous Localization and Mapping (SLAM) plays an important role in outdoor and indoor applications ranging from autonomous driving to indoor robotics. Outdoor SLAM has been widely used with the assistance of LiDAR or GPS. For indoor applications, the LiDAR technique does not satisfy the accuracy requirement and the GPS signals will be lost. An accurate and efficient scene sensing technique is required for indoor SLAM. As the most promising 3D sensing technique, the opportunities for indoor SLAM with fringe projection profilometry (FPP) systems are obvious, but methods to date have not fully leveraged the accuracy and speed of sensing that such systems offer. In this paper, we propose a novel FPP-based indoor SLAM method based on the coordinate transformation relationship of FPP, where the 2D-to-3D descriptor-assisted is used for mapping and localization. The correspondences generated by matching descriptors are used for fast and accurate mapping, and the transform estimation between the 2D and 3D descriptors is used to localize the sensor. The provided experimental results demonstrate that the proposed indoor SLAM can achieve the localization and mapping accuracy around one millimeter.
\end{abstract}

\end{frontmatter}
\linenumbers

\section{Introduction}
Simultaneous Localization and Mapping (SLAM) is of great importance in 3D computer vision with many applications in autonomous driving \cite{cadena2016past, bresson2017simultaneous}, indoor robotics \cite{azzam2020feature, hornung2010humanoid}, building surveying and mapping \cite{zeybek2021indoor, otero2020mobile}, etc. In general, SLAM can be divided into two categories of outdoor \cite{newman2006outdoor} and indoor SLAM \cite{zlatanova2013problems}. Outdoor SLAM applications such as autonomous driving and drone cruising work in relatively large scenes, and accordingly only decimeter (dm)- \cite{bresson2017simultaneous} or even meter (m)-level \cite{newman2006outdoor} SLAM accuracy is required. In contrast, indoor SLAM applications such as indoor robotics and building surveying \cite{zeybek2021indoor} usually work in narrow scenes including small scattered objects, and accordingly the higher SLAM accuracy is necessary such as millimeter (mm)-level, which is still unachievable for existing inddor SLAM technologies. 

       A typical SLAM system includes the front-end sensors that perceive surrounding the unknown scene data and the back-end algorithms for building a map of the scene and determining the {{sensors}’} location within this map \cite{durrant2006simultaneous}. The performance of SLAM usually relies on the employed sensors, i.e., the data perceived from the environment. Recently, the development of LiDAR and the assistance of GPS signals enables outdoor SLAM applications successfully solved \cite{chen2019probabilistic, ellis2016mapping}. However, indoor SLAM cannot receive external GPS signals, and its carrier is usually a small robot with low cost \cite{thrun1998learning}. Therefore, it is critical to seek a low-cost, small-volume, and high-accuracy sensor for indoor SLAM.

In SLAM, the 3D scenes are perceived by using the LiDAR \cite{ren2019robust}, stereo vision \cite{mur2017visual} or time of flight (ToF) \cite{newcombe2011kinectfusion} techniques. The LiDAR has been widely used in outdoor SLAM for autonomous driving due to its advantages of large working distance and high robustness. However, indoor SLAM using LiDAR is a challenging task due to the sparse nature of the obtained 3D data and computational complexities, which is difficult to provide real-time dense maps for indoor service robots \cite{li2016feature}. Alternatively, the stereo vision uses two cameras to simulate the human eye according to the principle of bionics and uses images collected from different angles to reconstruct 3D data. However, the shortcomings of stereo vision in terms of accuracy, stability, and field of view limit its application \cite{eade2006scalable, mustafah2012indoor}. The ToF technique obtains target depth by probing the round-trip time of light emitted and received by each pixel on the imaging chip, which has been widely used in commercial RGB-D sensors (e.g., Microsoft Kinect) \cite{newcombe2011kinectfusion}.  However, it has defects such as a difficult trade-off between the miniaturization and high pixel resolution of the imaging chip, large measurement errors in the near range, and slow measurement speed. As aforementioned, the exist 3D sensors used in SLAM do not satisfy the requirements of indoor millimeter (mm)-level SLAM, i.e., localizing the sensor and the interaction with the indoor scene. 

Optical metrology has been widely used for accurate and flexible 3D perceiving. Among plenty of state-of-art techniques, fringe projection profilometry (FPP) can achieve high-accuracy (i.e. micrometer-level) and high-speed (i.e. thousands of fps) indoor scene perceiving because the sub-pixel accurate camera-projector correspondences and binary defocus technology \cite{zhang2010recent, zheng2020high}. Compared with other techniques, FPP also has the inherent advantages of simple hardware configuration, flexibility in implementation, and dense point clouds acquisition \cite{gorthi2010fringe}. It is appealing to utilize the FPP sensor to achieve indoor SLAM. However, there are two challenges for the application of FPP sensor to indoor SLAM. First, a fast and robust multi-views registration is required for FPP to sense and map a large scene. Second, estimating camera localization based on FPP coordinate system and subsequent localization optimization is still in the blank. 

In order to tackle these challenges, in this paper, we propose a FPP-based indoor SLAM method. First, the FPP technique is introduced to perceive the 3D indoor scenes. Based on the perceived data of FPP, the correspond algorithm is proposed for millimeter (mm)-level indoor SLAM, which utilizes the 2D-to-3D descriptor and the coordinate transformation relationship of FPP. The mapping is achieved by solving the transformation matrix based on the correspondences generated by matching descriptors. The localization of sensor is achieved by transformation estimation between the 2D and 3D descriptors. With our method, users or robots can simply obtain accurate indoor maps and immediate feedback during perceiving. We also show both qualitative and quantitative results relating to various aspects of our FPP-based indoor method. 
The rest of this paper is organized as follows. Section 2 presents the principle of the proposed method. Section 3 provides experiments. Section 4 concludes this paper.

\section{Principle}

The flowchart of the proposed FPP-based indoor SLAM method is provided in Fig. 1, which includes the front-end using FPP technique to perceive the unknown scene, and the correspond back-end algorithm using the 2D-to-3D descriptor and the coordinate transformation relationship of FPP to estimate of the pose of sensors and construct the map of indoor scene. Specifically, for the back-end, the local mapping module is performed to generate local maps by minimizing errors between the transformed and original point clouds. The corresponding location and the pose of the sensor can be calculated by utilizing the coordinate-map given by FPP. The optimizer then performs localization optimization to eliminate the cumulative errors and revise local mapping. Finally, the multi-view point clouds are aligned in global maps. 

\begin{figure}[htbp] 
 \center{\includegraphics[width=15.5cm]  {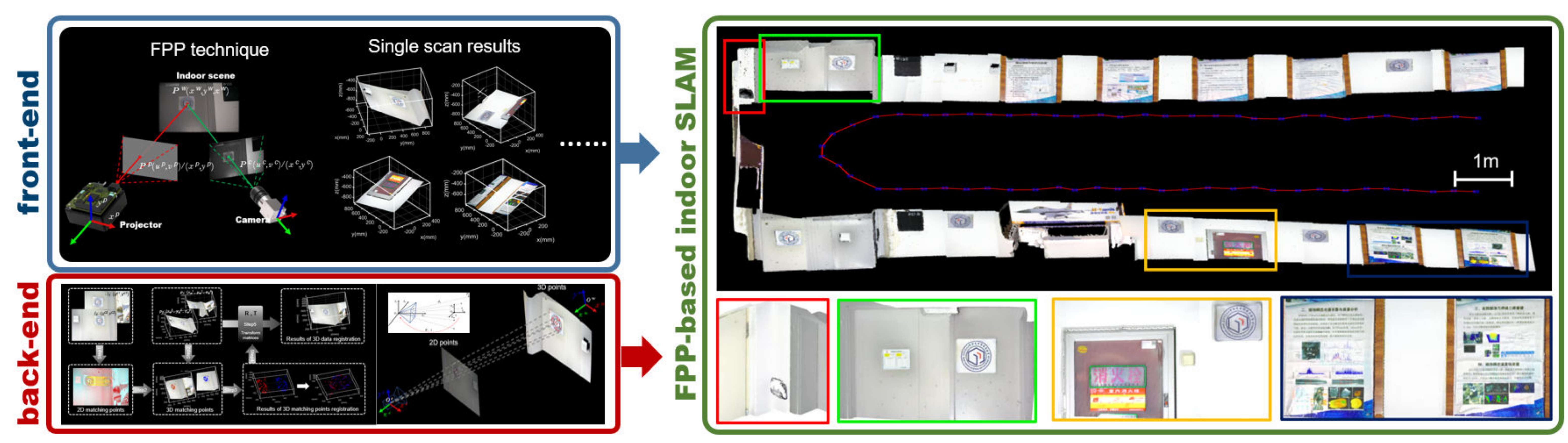}} 
 \caption{\label{1} Overall workflow of the proposed FPP-based SLAM method. } 
 \end{figure}

\subsection{The front-end: FPP technique}
The procedure of the FPP technique is provided in Fig. 2. A typical FPP system usually consists of a projector and a camera. The former is used to project coded fringe patterns onto the measured object, and the latter is used to synchronously capture the height-modulated fringe patterns. The projector can be seen as an inverse camera and the patterns help to establish distinct and accurate camera-projector correspondences regardless of whether the surface is textured or not. The desired phase can be calculated from these captured fringe patterns by using the transform-based \cite{su2001fourier} or phase-shifting algorithms \cite{huang2010comparison}. The calculated phase is always wrapped in a range of $\left( { - \pi ,\pi } \right]$ , and a phase unwrapping process is necessary to obtain the absolute phase \cite{zheng2020high}. The image correspondence between the projector and the camera can be established from the absolute phase \cite{zuo2016temporal}. The 3-D shape can be reconstructed from the image correspondence by using the triangulation method, when combining the image correspondence with system parameters calibrated before \cite{zhang2006novel}. 

\begin{figure}[htbp] 
 \center{\includegraphics[width=15.5cm]  {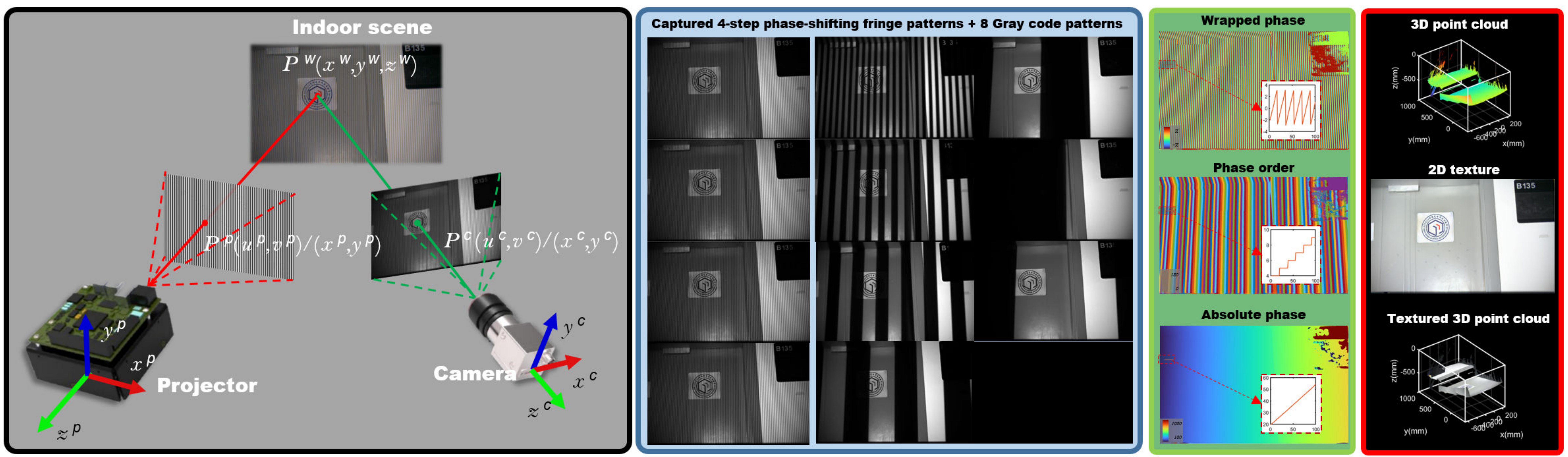}} 
 \caption{\label{1} Schematic diagram of FPP. } 
 \end{figure}

As aforementioned, both the system calibration and 3-D reconstruction rely on the retrieved phase, i.e., the wrapped phase and the absolute phase. The phase-shifting algorithm is preferred, and the captured sinusoidal fringe patterns can be described by

\begin{equation} \label{eq1}
    I_n(u^c,v^c)=a(u^c,v^c)+b(u^c,v^c)cos[\varphi(u^c,v^c)-\delta_n],n=1,2,...N,
\end{equation}
where $(u^c,v^c)$ denotes the pixel coordinate, $N$ denotes phase steps, $\delta_n$=$2\pi(n-1)/N$ denotes the phase shift amount, $a$, $b$ and $\varphi$ denote the background, fringe amplitude and desire phase, respecitively. The desire phase can be calculated by using a least-squares algorithm

\begin{equation} \label{eq2}
    {\varphi}(u^c,v^c)=arctan\left\{{-}\frac{\Sigma^N_{n=1}[I_n(u^c,v^c)sin(\delta_n)]}{\Sigma^N_{n=1}[I_n(u^c,v^c)cos(\delta_n)]}\right\}.
\end{equation}

To obtain the correct absolute phase, the Gray code-based phase unwrapping method is preferred due to its high robustness \cite{zhang2018absolute}, and a set of Gray code-based patterns is used to determine the fringe order of each fringe period. The absolute phase can be obtained by \cite{zheng2017phase}

\begin{equation} \label{eq3}
    \Phi(u^c,v^c)=\varphi(u^c,v^c)+K(u^c,v^c)\times2\pi,
\end{equation} 
where $K$ denotes the determined fringe order.

In this paper, both the projector and the camera can be regarded as a classical hole model and the inverse-camera-based method is applied to reconstruct the 3D points from the absolute phase. The pinhole camera model maps 3D points in the world coordinate frame $(x^w, y^w, z^w)$ to 2D pixel coordinates $(u, v)$ . This gives the equations

\begin{equation} \label{eq4}
\begin{aligned}
{s^c}\left[ \begin{array}{l}
{u^c}\\
{v^c}\\
1
\end{array} \right] = {A^c}\left[ \begin{array}{l}
{x^w}\\
{y^w}\\
{z^w}\\
1
\end{array} \right] = \left[ \begin{array}{l}
a_{11}^c{\kern 1pt} {\kern 1pt} {\kern 1pt} {\kern 1pt} {\kern 1pt} {\kern 1pt} {\kern 1pt} a_{12}^c{\kern 1pt} {\kern 1pt} {\kern 1pt} {\kern 1pt} {\kern 1pt} {\kern 1pt} {\kern 1pt} a_{13}^c{\kern 1pt} {\kern 1pt} {\kern 1pt} {\kern 1pt} {\kern 1pt} {\kern 1pt} {\kern 1pt} a_{14}^c{\kern 1pt} \\
a_{21}^c{\kern 1pt} {\kern 1pt} {\kern 1pt} {\kern 1pt} {\kern 1pt} {\kern 1pt} {\kern 1pt} a_{22}^c{\kern 1pt} {\kern 1pt} {\kern 1pt} {\kern 1pt} {\kern 1pt} {\kern 1pt} {\kern 1pt} a_{23}^c{\kern 1pt} {\kern 1pt} {\kern 1pt} {\kern 1pt} {\kern 1pt} {\kern 1pt} {\kern 1pt} a_{24}^c\\
a_{31}^c{\kern 1pt} {\kern 1pt} {\kern 1pt} {\kern 1pt} {\kern 1pt} {\kern 1pt} {\kern 1pt} a_{32}^c{\kern 1pt} {\kern 1pt} {\kern 1pt} {\kern 1pt} {\kern 1pt} {\kern 1pt} {\kern 1pt} a_{33}^c{\kern 1pt} {\kern 1pt} {\kern 1pt} {\kern 1pt} {\kern 1pt} {\kern 1pt} {\kern 1pt} a_{34}^c
\end{array} \right]\left[ \begin{array}{l}
{x^w}\\
{y^w}\\
{z^w}\\
1
\end{array} \right],
\end{aligned}
\end{equation}

\begin{equation} \label{eq5}
\begin{aligned}
{s^p}\left[ \begin{array}{l}
{u^p}\\
{v^p}\\
1
\end{array} \right] = {A^p}\left[ \begin{array}{l}
{x_w}\\
{y_w}\\
{z_w}\\
1
\end{array} \right] = \left[ \begin{array}{l}
a_{11}^p{\kern 1pt} {\kern 1pt} {\kern 1pt} {\kern 1pt} {\kern 1pt} {\kern 1pt} {\kern 1pt} a_{12}^p{\kern 1pt} {\kern 1pt} {\kern 1pt} {\kern 1pt} {\kern 1pt} {\kern 1pt} {\kern 1pt} a_{13}^p{\kern 1pt} {\kern 1pt} {\kern 1pt} {\kern 1pt} {\kern 1pt} {\kern 1pt} {\kern 1pt} a_{14}^p{\kern 1pt} \\
a_{21}^p{\kern 1pt} {\kern 1pt} {\kern 1pt} {\kern 1pt} {\kern 1pt} {\kern 1pt} {\kern 1pt} a_{22}^p{\kern 1pt} {\kern 1pt} {\kern 1pt} {\kern 1pt} {\kern 1pt} {\kern 1pt} {\kern 1pt} a_{23}^p{\kern 1pt} {\kern 1pt} {\kern 1pt} {\kern 1pt} {\kern 1pt} {\kern 1pt} {\kern 1pt} a_{24}^p\\
a_{31}^p{\kern 1pt} {\kern 1pt} {\kern 1pt} {\kern 1pt} {\kern 1pt} {\kern 1pt} {\kern 1pt} a_{32}^p{\kern 1pt} {\kern 1pt} {\kern 1pt} {\kern 1pt} {\kern 1pt} {\kern 1pt} {\kern 1pt} a_{33}^p{\kern 1pt} {\kern 1pt} {\kern 1pt} {\kern 1pt} {\kern 1pt} {\kern 1pt} {\kern 1pt} a_{34}^p
\end{array} \right]\left[ \begin{array}{l}
{x^w}\\
{y^w}\\
{z^w}\\
1
\end{array} \right],
\end{aligned}
\end{equation}

where superscript $c/p$  represents camera/projector and $s$ denotes the scaling factor. $A$ of $3 \times 4$ matrix is the product of the intrinsic and extrinsic matrix, which is expressed as 

\begin{equation} \label{eq6}
\begin{aligned}
{A^c} = {I^c} \times {\kern 1pt} {\kern 1pt} {\kern 1pt} {\kern 1pt} [{R^c}{\kern 1pt} |{\kern 1pt} {\kern 1pt} {T^c}] = \left[ \begin{array}{l}
f_x^c{\kern 1pt} {\kern 1pt} {\kern 1pt} {\kern 1pt} {\kern 1pt} {\kern 1pt} {\kern 1pt} 0{\kern 1pt} {\kern 1pt} {\kern 1pt} {\kern 1pt} {\kern 1pt} {\kern 1pt} {\kern 1pt} {\kern 1pt} {\kern 1pt} {\kern 1pt} {\kern 1pt} {\kern 1pt} u_0^c{\kern 1pt} \\
0{\kern 1pt} {\kern 1pt} {\kern 1pt} {\kern 1pt} {\kern 1pt} {\kern 1pt} {\kern 1pt} {\kern 1pt} {\kern 1pt} {\kern 1pt} {\kern 1pt} f_y^c{\kern 1pt} {\kern 1pt} {\kern 1pt} {\kern 1pt} {\kern 1pt} {\kern 1pt} {\kern 1pt} v_0^c{\kern 1pt} \\
0{\kern 1pt} {\kern 1pt} {\kern 1pt} {\kern 1pt} {\kern 1pt} {\kern 1pt} {\kern 1pt} {\kern 1pt} {\kern 1pt} {\kern 1pt} {\kern 1pt} {\kern 1pt} {\kern 1pt} {\kern 1pt} 0{\kern 1pt} {\kern 1pt} {\kern 1pt} {\kern 1pt} {\kern 1pt} {\kern 1pt} {\kern 1pt} {\kern 1pt} {\kern 1pt} {\kern 1pt} {\kern 1pt} {\kern 1pt} {\kern 1pt} {\kern 1pt} 1{\kern 1pt} 
\end{array} \right]\left[ \begin{array}{l}
r_{11}^c{\kern 1pt} {\kern 1pt} {\kern 1pt} {\kern 1pt} {\kern 1pt} {\kern 1pt} {\kern 1pt} r_{12}^c{\kern 1pt} {\kern 1pt} {\kern 1pt} {\kern 1pt} {\kern 1pt} {\kern 1pt} {\kern 1pt} r_{13}^c{\kern 1pt} {\kern 1pt} {\kern 1pt} {\kern 1pt} {\kern 1pt} {\kern 1pt} {\kern 1pt} t_1^c{\kern 1pt} \\
r_{21}^c{\kern 1pt} {\kern 1pt} {\kern 1pt} {\kern 1pt} {\kern 1pt} {\kern 1pt} {\kern 1pt} r_{22}^c{\kern 1pt} {\kern 1pt} {\kern 1pt} {\kern 1pt} {\kern 1pt} {\kern 1pt} {\kern 1pt} r_{23}^c{\kern 1pt} {\kern 1pt} {\kern 1pt} {\kern 1pt} {\kern 1pt} {\kern 1pt} {\kern 1pt} t_2^c\\
r_{31}^c{\kern 1pt} {\kern 1pt} {\kern 1pt} {\kern 1pt} {\kern 1pt} {\kern 1pt} {\kern 1pt} r_{32}^c{\kern 1pt} {\kern 1pt} {\kern 1pt} {\kern 1pt} {\kern 1pt} {\kern 1pt} {\kern 1pt} r_{33}^c{\kern 1pt} {\kern 1pt} {\kern 1pt} {\kern 1pt} {\kern 1pt} {\kern 1pt} {\kern 1pt} t_3^c
\end{array} \right],
\end{aligned}
\end{equation}

\begin{equation} \label{eq7}
\begin{aligned}
{A^p} = {I^p} \times {\kern 1pt} {\kern 1pt} {\kern 1pt} {\kern 1pt} [{R^p}{\kern 1pt} {\kern 1pt} |{\kern 1pt} {T^p}] = \left[ \begin{array}{l}
f_x^p{\kern 1pt} {\kern 1pt} {\kern 1pt} {\kern 1pt} {\kern 1pt} {\kern 1pt} {\kern 1pt} 0{\kern 1pt} {\kern 1pt} {\kern 1pt} {\kern 1pt} {\kern 1pt} {\kern 1pt} {\kern 1pt} {\kern 1pt} {\kern 1pt} {\kern 1pt} {\kern 1pt} {\kern 1pt} u_0^p{\kern 1pt} \\
0{\kern 1pt} {\kern 1pt} {\kern 1pt} {\kern 1pt} {\kern 1pt} {\kern 1pt} {\kern 1pt} {\kern 1pt} {\kern 1pt} {\kern 1pt} {\kern 1pt} f_y^p{\kern 1pt} {\kern 1pt} {\kern 1pt} {\kern 1pt} {\kern 1pt} {\kern 1pt} {\kern 1pt} v_0^p{\kern 1pt} \\
0{\kern 1pt} {\kern 1pt} {\kern 1pt} {\kern 1pt} {\kern 1pt} {\kern 1pt} {\kern 1pt} {\kern 1pt} {\kern 1pt} {\kern 1pt} {\kern 1pt} {\kern 1pt} {\kern 1pt} {\kern 1pt} 0{\kern 1pt} {\kern 1pt} {\kern 1pt} {\kern 1pt} {\kern 1pt} {\kern 1pt} {\kern 1pt} {\kern 1pt} {\kern 1pt} {\kern 1pt} {\kern 1pt} {\kern 1pt} {\kern 1pt} {\kern 1pt} 1{\kern 1pt} 
\end{array} \right]\left[ \begin{array}{l}
r_{11}^p{\kern 1pt} {\kern 1pt} {\kern 1pt} {\kern 1pt} {\kern 1pt} {\kern 1pt} {\kern 1pt} r_{12}^p{\kern 1pt} {\kern 1pt} {\kern 1pt} {\kern 1pt} {\kern 1pt} {\kern 1pt} {\kern 1pt} r_{13}^p{\kern 1pt} {\kern 1pt} {\kern 1pt} {\kern 1pt} {\kern 1pt} {\kern 1pt} {\kern 1pt} t_1^p{\kern 1pt} \\
r_{21}^p{\kern 1pt} {\kern 1pt} {\kern 1pt} {\kern 1pt} {\kern 1pt} {\kern 1pt} {\kern 1pt} r_{22}^p{\kern 1pt} {\kern 1pt} {\kern 1pt} {\kern 1pt} {\kern 1pt} {\kern 1pt} {\kern 1pt} r_{23}^p{\kern 1pt} {\kern 1pt} {\kern 1pt} {\kern 1pt} {\kern 1pt} {\kern 1pt} {\kern 1pt} t_2^p\\
r_{31}^p{\kern 1pt} {\kern 1pt} {\kern 1pt} {\kern 1pt} {\kern 1pt} {\kern 1pt} {\kern 1pt} r_{32}^p{\kern 1pt} {\kern 1pt} {\kern 1pt} {\kern 1pt} {\kern 1pt} {\kern 1pt} {\kern 1pt} r_{33}^p{\kern 1pt} {\kern 1pt} {\kern 1pt} {\kern 1pt} {\kern 1pt} {\kern 1pt} {\kern 1pt} t_3^p
\end{array} \right],
\end{aligned}
\end{equation}

$I$ is the $3 \times 3$ intrinsic parameter, $f_x$ and $f_y$ are the projector focal, and $(u_0,v_0)$ is the principal coordinate. $[R | T]$ is known as the extrinsic matrix. $R$ denotes the $3 \times 3$  rotational matric and $T \in R^3$ denotes the translation vector that describes the orientation and position of the world frame. 

When the FPP system is calibrated by capturing patterns on a planar surface at several positions, $A^c$ and $A^p$ become the known values. Next, when the corresponding point between the camera and projector is matched, the 3D coordinate can be solved. In the FPP system, the absolute phase can satisfy the matching requirement. For example, the value of the transverse absolute phase of the point ${p^c}({u^c},{v^c})$ on the camera is ${\Phi ^c}({u^c},{v^c})$ , and $u^p$ of the according to the point $p^p$ on the projector can be matched by Eq. (8).

\begin{equation} \label{eq8}
\begin{aligned}
{u^p} = \frac{{{\Phi ^c}({u^c},{v^c}) \times \lambda }}{{2\pi }},
\end{aligned}
\end{equation}

where ${\Phi ^c}({u^c},{v^c})$ represents the transverse absolute phase, and $\lambda$ is the fringe width. If ${v^p}$ is required, the vertical absolute phase can be used. When ${u^p}$ is matched, the 3D coordinate $({x^w},{y^w},{z^w})$ can be obtained by the following equation.

\begin{equation} \label{eq9}
\begin{aligned}
\left[ \begin{array}{l}
{x^w}\\
{y^w}\\
{z^w}
\end{array} \right] = {\left[ \begin{array}{l}
a_{11}^c - {u^c}a_{31}^c{\kern 1pt} {\kern 1pt} {\kern 1pt} {\kern 1pt} {\kern 1pt} {\kern 1pt} {\kern 1pt} a_{12}^c - {u^c}a_{32}^c{\kern 1pt} {\kern 1pt} {\kern 1pt} {\kern 1pt} {\kern 1pt} {\kern 1pt} {\kern 1pt} a_{13}^c - {u^c}a_{33}^c{\kern 1pt} \\
a_{21}^c - {v^c}a_{31}^c{\kern 1pt} {\kern 1pt} {\kern 1pt} {\kern 1pt} {\kern 1pt} {\kern 1pt} {\kern 1pt} a_{22}^c - {v^c}a_{32}^c{\kern 1pt} {\kern 1pt} {\kern 1pt} {\kern 1pt} {\kern 1pt} {\kern 1pt} {\kern 1pt} a_{23}^c - {v^c}a_{33}^c\\
a_{31}^c{\kern 1pt} {\kern 1pt}  - {u^p}a_{31}^c{\kern 1pt} {\kern 1pt} {\kern 1pt} {\kern 1pt} {\kern 1pt} a_{32}^c - {u^p}a_{31}^c{\kern 1pt} {\kern 1pt} {\kern 1pt} {\kern 1pt} {\kern 1pt} {\kern 1pt} {\kern 1pt} a_{33}^c - {u^p}a_{31}^c{\kern 1pt} 
\end{array} \right]^{ - 1}}\left[ \begin{array}{l}
{u^c}a_{34}^c - {\kern 1pt} a_{14}^c\\
{v^c}a_{34}^c - a_{24}^c\\
{u^p}a_{34}^c{\kern 1pt} {\kern 1pt}  - a_{24}^c
\end{array} \right],
\end{aligned}
\end{equation}
Finally, the 2D texture images are rendered point-to-point onto the point clouds.

\subsection{2.2	The back-end:}

\subsubsection{Basics of mapping}
After obtaining high-quality data, a coarse-to-fine pairwise registration algorithm based on the 2D-to-3D descriptor is proposed for fast and accurate mapping, which is achieved by solving the transformation matrix based on the correspondences generated by matching descriptors. Specifically, we apply the coordinate transformation to convert the affine transformation between 2D feature points to a 3D transformation to provide an initial prior for the corresponding point clouds registration. Then, in order to improve the efficiency of registration, 3D feature points are used instead of 3D point clouds to perform registration. The entire process of mapping method is shown in Fig. 3. Step 1: extract 2D matching points with SURF algorithm and obtain 2D transformation matrix; Step 2: convert the 2D transformation matrix into the 3D transformation matrix according to the coordinate transformation, and use it as the initial prior for registration; Step 3: extract the corresponding 3D points according to the 2D feature points; Step 4: perform ICP on the 3D points to obtain the transformation matrix combined with the initial prior; Step 5: apply the transformation matrix to perform registration on the point clouds; Step 6: return to Step 1 and repeat the above processes.

\begin{figure}[htbp] 
 \center{\includegraphics[width=10cm]  {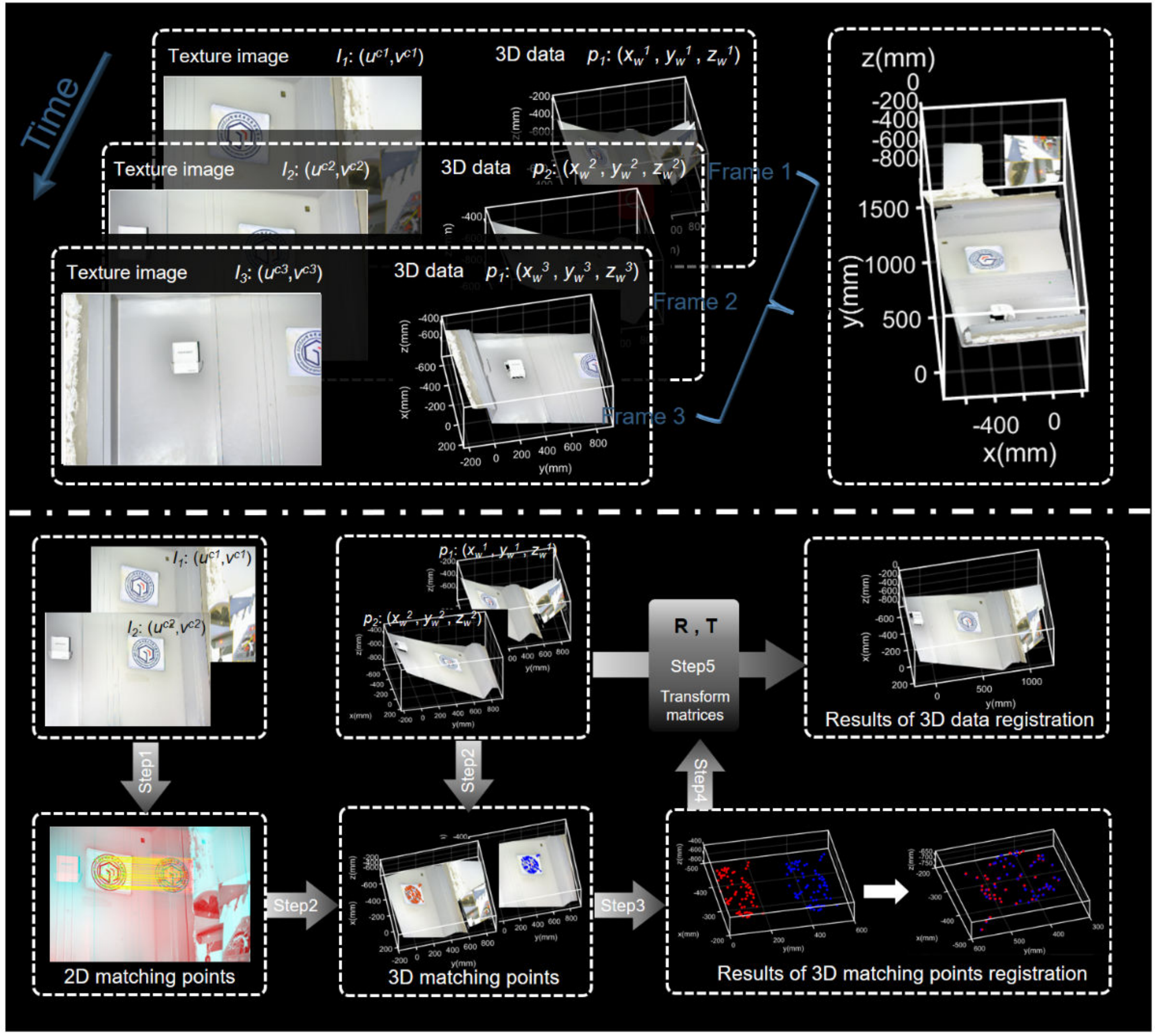}} 
 \caption{\label{1} Overall workflow of the proposed FPP-based SLAM method. } 
 \end{figure}

Define two adjacent 3D frames as Frame 1 and Frame 2, which have corresponding 2D texture images ${I_1}:(u_c^1,v_c^1)$ and ${I_2}:(u_c^2,v_c^2)$, and 3D data $p1:(x_w^1,y_w^1,z_w^1)$ and $p2:(x_w^2,y_w^2,z_w^2)$ . First, the key points   and   are obtained and matched by Speeded Up Robust Features (SURF) \cite{triggs1999bundle} algorithm. In order to improve the registration efficiency, we use paired 3D feature points instead of complete 3D point clouds for registration. By using Eq. 9, we record sets of 3D matching points corresponding to the found 2D ones as   and  . The corresponding 3D points can be used to calculate the rotation matrix R and initial translation vector T of the coordinate transformation. The calculation can be formulated as

\begin{equation} \label{eq10}
\begin{aligned}
{min}\sum_{i=1}^n \left \|{{P_1}^{'} - {RP_2}^{'} - T}\right\|^2,
\end{aligned}
\end{equation}

The obtained transformation of 2D matching points can be transformed to 3D transformation to initialize the iteration. After ${P_1}^{'}$ and ${RP_2}^{'}$ are roughly registered, they almost have been registered well. To improve the registration accuracy, the result is further refined by using the ICP algorithm \cite{rusinkiewicz2001efficient}. Then, for stitching a sequence of point clouds, the next point cloud data can be as the target data and the target data registered in the previous step can be as the source data. Repeating the above two frame point clouds registration process to achieve a series of point cloud registration.

\subsubsection{Basics of localization}
Live camera localization involves estimating the current camera position and pose for each new point data, which is achieved by transformation estimation between the 2D and 3D descriptors. The 3D-2D points matching is performed to estimate the pose of the sensor using the 3D date of the perceived indoor scene. Specifically, we find the correspondences between camera and world coordinate points, and calculate the camera pose through optimization.

The schematic of the camera localization is shown in  Fig. 4, the transformation relationship between the world coordinate system $O_w$ and the camera coordinate system $O_c$ can be simply described as

\begin{figure}[htbp] 
 \center{\includegraphics[width=8.5cm]  {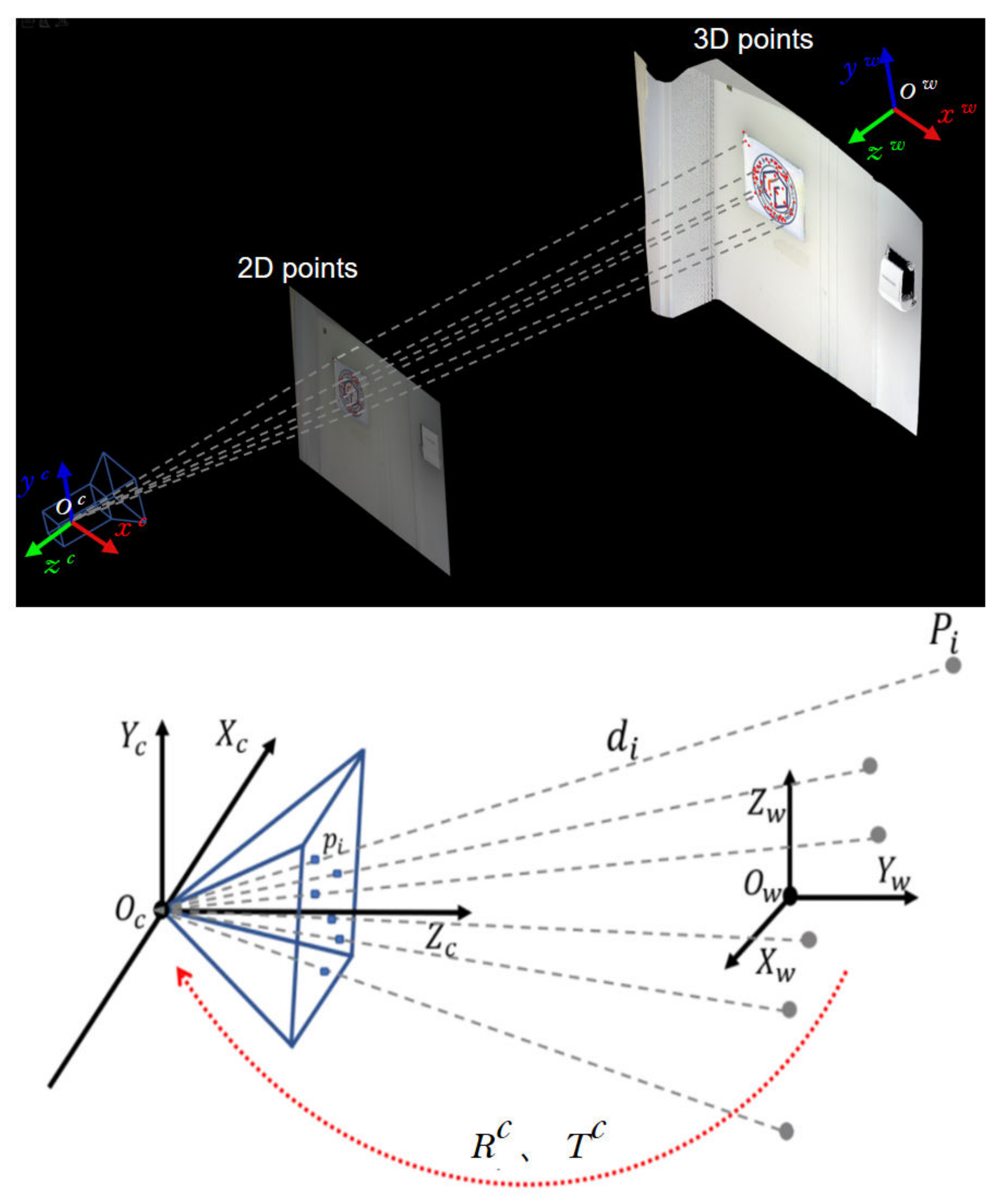}} 
 \caption{\label{1} Overall workflow of the proposed FPP-based SLAM method. } 
 \end{figure}

\begin{equation} \label{eq11}
\begin{aligned}
{O_c} = {R^c}{O_w} + {T^c}.
\end{aligned}
\end{equation}

Thus, the formula can be transformed into Eq. 12.

\begin{equation} \label{eq12}
\begin{aligned}
{O_w} = {({R^c})^{ - 1}}{O_c} - {({R^c})^{ - 1}}{T^c}.
\end{aligned}
\end{equation}

Because the camera localization is for the optical center coordinates, and the optical center coordinates are the origin   relative to the camera coordinates. Considering   is an orthogonal matrix, so ${({R^c})^T} = {({R^c})^{ - 1}}$. 

\begin{equation} \label{eq13}
\begin{aligned}
{O_w} =  - {({R^c})^{ - 1}}{T^c} =  - {({R^c})^T}{T^c},
\end{aligned}
\end{equation}
Thus, if we can calculate the rotation matrix and translation matrix of the current captured point data, the current camera position and pose can be easily obtained.

For the FPP imaging system, the coordinates of the 3D point cloud in real space are $(x^w,y^w,z^w)$, and the homogeneous coordinates can be expressed as $(x^w,y^w,z^w,1)$. The corresponding projected points in the camera coordinate are $(u^c,v^c)$, and the homogeneous coordinates can be expressed as $(u^c,v^c,1)$. The camera's internal parameter $I^c$  has been obtained through device calibration. By using Eq.8 and Eq.9, we can derive the following form.

\begin{equation} \label{eq14}
\begin{aligned}
\begin{array}{l}
s_{_{}}^cu_{_{}}^c = a_{11}^cx_{}^w + a_{12}^cy_{}^w + a_{13}^cz_{}^w + a_{14}^c\\
s_{_{}}^cv_{_{}}^c = a_{21}^cx_{}^w + a_{22}^cy_{}^w + a_{23}^cz_{}^w + a_{24}^c\\
s_{_{}}^c = a_{31}^cx_{}^w + a_{32}^cy_{}^w + a_{33}^cz_{}^w + a_{34}^c
\end{array},
\end{aligned}
\end{equation}

By eliminating $s^c$, Eq. 14 can be simplified as 

\begin{equation} \label{eq15}
\begin{aligned}
\begin{array}{l}
a_{11}^cx_{_{}}^w + a_{12}^cy_{_{}}^w + a_{13}^cz_{_{}}^w + a_{14}^c - a_{31}^cx_{_{}}^wu_{_{}}^c - a_{32}^cy_{_{}}^wu_{_{}}^c - a_{33}^cz_{_{}}^wu_{_{}}^c - a_{34}^cu_{_{}}^c = 0\\
a_{21}^cx_{_{}}^w + a_{22}^cy_{_{}}^w + a_{23}^cz_{_{}}^w + a_{24}^c - a_{31}^cx_{_{}}^wv_{_{}}^c - a_{32}^cy_{_{}}^wv_{_{}}^c - a_{33}^cz_{_{}}^wv_{_{}}^c - a_{34}^cv_{_{}}^c = 0
\end{array},
\end{aligned}
\end{equation}
According to the above formula, each set of 3D-2D matching points corresponds to two equations, there are a total of 12 unknowns, and at least 6 sets of matching points are required.

Assuming there are $N$ sets of matching points, Eq.15 can be rewritten as 

\begin{equation} \label{eq16}
\begin{aligned}
\left[ \begin{array}{l}
x_1^w{\kern 1pt} {\kern 1pt} {\kern 1pt} {\kern 1pt} {\kern 1pt} {\kern 1pt} {\kern 1pt} {\kern 1pt} {\kern 1pt} y_1^w{\kern 1pt} {\kern 1pt} {\kern 1pt} {\kern 1pt} {\kern 1pt} {\kern 1pt} {\kern 1pt} {\kern 1pt} {\kern 1pt} z_1^w{\kern 1pt} {\kern 1pt} {\kern 1pt} {\kern 1pt} {\kern 1pt} {\kern 1pt} {\kern 1pt} {\kern 1pt} {\kern 1pt} {\kern 1pt} {\kern 1pt} {\kern 1pt} 1{\kern 1pt} {\kern 1pt} {\kern 1pt} {\kern 1pt} {\kern 1pt} {\kern 1pt} {\kern 1pt} {\kern 1pt} {\kern 1pt} {\kern 1pt} {\kern 1pt} {\kern 1pt} {\kern 1pt} {\kern 1pt} {\kern 1pt} 0{\kern 1pt} {\kern 1pt} {\kern 1pt} {\kern 1pt} {\kern 1pt} {\kern 1pt} {\kern 1pt} {\kern 1pt} {\kern 1pt} {\kern 1pt} {\kern 1pt} {\kern 1pt} {\kern 1pt} {\kern 1pt} {\kern 1pt} {\kern 1pt} {\kern 1pt} {\kern 1pt} 0{\kern 1pt} {\kern 1pt} {\kern 1pt} {\kern 1pt} {\kern 1pt} {\kern 1pt} {\kern 1pt} {\kern 1pt} {\kern 1pt} {\kern 1pt} {\kern 1pt} {\kern 1pt} {\kern 1pt} {\kern 1pt} {\kern 1pt} {\kern 1pt} 0{\kern 1pt} {\kern 1pt} {\kern 1pt} {\kern 1pt} {\kern 1pt} {\kern 1pt} {\kern 1pt} {\kern 1pt} {\kern 1pt} {\kern 1pt} {\kern 1pt} {\kern 1pt} {\kern 1pt} {\kern 1pt} 0{\kern 1pt} {\kern 1pt} {\kern 1pt} {\kern 1pt} {\kern 1pt} {\kern 1pt} {\kern 1pt} {\kern 1pt} {\kern 1pt} {\kern 1pt} {\kern 1pt} {\kern 1pt}  - u_1^cx_1^w{\kern 1pt} {\kern 1pt} {\kern 1pt} {\kern 1pt} {\kern 1pt} {\kern 1pt} {\kern 1pt} {\kern 1pt} {\kern 1pt} {\kern 1pt} {\kern 1pt} {\kern 1pt} {\kern 1pt}  - u_1^cy_1^w{\kern 1pt} {\kern 1pt} {\kern 1pt} {\kern 1pt} {\kern 1pt} {\kern 1pt} {\kern 1pt} {\kern 1pt} {\kern 1pt} {\kern 1pt} {\kern 1pt} {\kern 1pt} {\kern 1pt}  - u_1^cz_1^w{\kern 1pt} {\kern 1pt} {\kern 1pt} {\kern 1pt} {\kern 1pt} {\kern 1pt} {\kern 1pt} {\kern 1pt} {\kern 1pt} {\kern 1pt} {\kern 1pt} {\kern 1pt} {\kern 1pt}  - u_1^c\\
0{\kern 1pt} {\kern 1pt} {\kern 1pt} {\kern 1pt} {\kern 1pt} {\kern 1pt} {\kern 1pt} {\kern 1pt} {\kern 1pt} {\kern 1pt} {\kern 1pt} {\kern 1pt} {\kern 1pt} {\kern 1pt} {\kern 1pt} {\kern 1pt} {\kern 1pt} 0{\kern 1pt} {\kern 1pt} {\kern 1pt} {\kern 1pt} {\kern 1pt} {\kern 1pt} {\kern 1pt} {\kern 1pt} {\kern 1pt} {\kern 1pt} {\kern 1pt} {\kern 1pt} {\kern 1pt} {\kern 1pt} 0{\kern 1pt} {\kern 1pt} {\kern 1pt} {\kern 1pt} {\kern 1pt} {\kern 1pt} {\kern 1pt} {\kern 1pt} {\kern 1pt} {\kern 1pt} {\kern 1pt} {\kern 1pt} {\kern 1pt} {\kern 1pt} {\kern 1pt} {\kern 1pt} 0{\kern 1pt} {\kern 1pt} {\kern 1pt} {\kern 1pt} {\kern 1pt} {\kern 1pt} {\kern 1pt} {\kern 1pt} {\kern 1pt} {\kern 1pt} {\kern 1pt} {\kern 1pt} x_1^w{\kern 1pt} {\kern 1pt} {\kern 1pt} {\kern 1pt} {\kern 1pt} {\kern 1pt} {\kern 1pt} {\kern 1pt} {\kern 1pt} {\kern 1pt} {\kern 1pt} y_1^w{\kern 1pt} {\kern 1pt} {\kern 1pt} {\kern 1pt} {\kern 1pt} {\kern 1pt} {\kern 1pt} {\kern 1pt} {\kern 1pt} {\kern 1pt} {\kern 1pt} z_1^w{\kern 1pt} {\kern 1pt} {\kern 1pt} {\kern 1pt} {\kern 1pt} {\kern 1pt} {\kern 1pt} {\kern 1pt} {\kern 1pt} {\kern 1pt} {\kern 1pt} 1{\kern 1pt} {\kern 1pt} {\kern 1pt} {\kern 1pt} {\kern 1pt} {\kern 1pt} {\kern 1pt} {\kern 1pt} {\kern 1pt} {\kern 1pt} {\kern 1pt} {\kern 1pt} {\kern 1pt}  - v_1^cx_1^w{\kern 1pt} {\kern 1pt} {\kern 1pt} {\kern 1pt} {\kern 1pt} {\kern 1pt} {\kern 1pt} {\kern 1pt} {\kern 1pt} {\kern 1pt} {\kern 1pt} {\kern 1pt} {\kern 1pt} {\kern 1pt}  - v_1^cy_1^w{\kern 1pt} {\kern 1pt} {\kern 1pt} {\kern 1pt} {\kern 1pt} {\kern 1pt} {\kern 1pt} {\kern 1pt} {\kern 1pt} {\kern 1pt} {\kern 1pt} {\kern 1pt} {\kern 1pt}  - v_1^cz_1^w{\kern 1pt} {\kern 1pt} {\kern 1pt} {\kern 1pt} {\kern 1pt} {\kern 1pt} {\kern 1pt} {\kern 1pt} {\kern 1pt} {\kern 1pt} {\kern 1pt} {\kern 1pt}  - v_1^c\\
...{\kern 1pt} {\kern 1pt} {\kern 1pt} {\kern 1pt} ...\\
x_N^w{\kern 1pt} {\kern 1pt} {\kern 1pt} {\kern 1pt} {\kern 1pt} {\kern 1pt} y_N^w{\kern 1pt} {\kern 1pt} {\kern 1pt} {\kern 1pt} {\kern 1pt} {\kern 1pt} {\kern 1pt} z_N^w{\kern 1pt} {\kern 1pt} {\kern 1pt} {\kern 1pt} {\kern 1pt} {\kern 1pt} {\kern 1pt} {\kern 1pt} {\kern 1pt} {\kern 1pt} {\kern 1pt} {\kern 1pt} {\kern 1pt} {\kern 1pt} {\kern 1pt} 1{\kern 1pt} {\kern 1pt} {\kern 1pt} {\kern 1pt} {\kern 1pt} {\kern 1pt} {\kern 1pt} {\kern 1pt} {\kern 1pt} {\kern 1pt} {\kern 1pt} {\kern 1pt} {\kern 1pt} {\kern 1pt} {\kern 1pt} 0{\kern 1pt} {\kern 1pt} {\kern 1pt} {\kern 1pt} {\kern 1pt} {\kern 1pt} {\kern 1pt} {\kern 1pt} {\kern 1pt} {\kern 1pt} {\kern 1pt} {\kern 1pt} {\kern 1pt} {\kern 1pt} {\kern 1pt} {\kern 1pt} {\kern 1pt} 0{\kern 1pt} {\kern 1pt} {\kern 1pt} {\kern 1pt} {\kern 1pt} {\kern 1pt} {\kern 1pt} {\kern 1pt} {\kern 1pt} {\kern 1pt} {\kern 1pt} {\kern 1pt} {\kern 1pt} {\kern 1pt} {\kern 1pt} {\kern 1pt} 0{\kern 1pt} {\kern 1pt} {\kern 1pt} {\kern 1pt} {\kern 1pt} {\kern 1pt} {\kern 1pt} {\kern 1pt} {\kern 1pt} {\kern 1pt} {\kern 1pt} {\kern 1pt} {\kern 1pt} {\kern 1pt} {\kern 1pt} 0{\kern 1pt} {\kern 1pt} {\kern 1pt} {\kern 1pt} {\kern 1pt} {\kern 1pt} {\kern 1pt} {\kern 1pt} {\kern 1pt} {\kern 1pt} {\kern 1pt}  - u_N^cx_N^w{\kern 1pt} {\kern 1pt} {\kern 1pt} {\kern 1pt} {\kern 1pt} {\kern 1pt} {\kern 1pt} {\kern 1pt} {\kern 1pt} {\kern 1pt}  - u_N^cy_N^w{\kern 1pt} {\kern 1pt} {\kern 1pt} {\kern 1pt} {\kern 1pt} {\kern 1pt} {\kern 1pt} {\kern 1pt} {\kern 1pt} {\kern 1pt} {\kern 1pt}  - u_N^cz_N^w{\kern 1pt} {\kern 1pt} {\kern 1pt} {\kern 1pt} {\kern 1pt} {\kern 1pt} {\kern 1pt} {\kern 1pt} {\kern 1pt}  - u_N^c\\
0{\kern 1pt} {\kern 1pt} {\kern 1pt} {\kern 1pt} {\kern 1pt} {\kern 1pt} {\kern 1pt} {\kern 1pt} {\kern 1pt} {\kern 1pt} {\kern 1pt} {\kern 1pt} {\kern 1pt} {\kern 1pt} {\kern 1pt} {\kern 1pt} 0{\kern 1pt} {\kern 1pt} {\kern 1pt} {\kern 1pt} {\kern 1pt} {\kern 1pt} {\kern 1pt} {\kern 1pt} {\kern 1pt} {\kern 1pt} {\kern 1pt} {\kern 1pt} 0{\kern 1pt} {\kern 1pt} {\kern 1pt} {\kern 1pt} {\kern 1pt} {\kern 1pt} {\kern 1pt} {\kern 1pt} {\kern 1pt} {\kern 1pt} {\kern 1pt} {\kern 1pt} {\kern 1pt} {\kern 1pt} {\kern 1pt} {\kern 1pt} {\kern 1pt} {\kern 1pt} {\kern 1pt} {\kern 1pt} 0{\kern 1pt} {\kern 1pt} {\kern 1pt} {\kern 1pt} {\kern 1pt} {\kern 1pt} {\kern 1pt} {\kern 1pt} {\kern 1pt} {\kern 1pt} {\kern 1pt} x_N^w{\kern 1pt} {\kern 1pt} {\kern 1pt} {\kern 1pt} {\kern 1pt} {\kern 1pt} {\kern 1pt} {\kern 1pt} {\kern 1pt} y_N^w{\kern 1pt} {\kern 1pt} {\kern 1pt} {\kern 1pt} {\kern 1pt} {\kern 1pt} {\kern 1pt} {\kern 1pt} {\kern 1pt} z_N^w{\kern 1pt} {\kern 1pt} {\kern 1pt} {\kern 1pt} {\kern 1pt} {\kern 1pt} {\kern 1pt} {\kern 1pt} {\kern 1pt} {\kern 1pt} {\kern 1pt} {\kern 1pt} {\kern 1pt} {\kern 1pt} 1{\kern 1pt} {\kern 1pt} {\kern 1pt} {\kern 1pt} {\kern 1pt} {\kern 1pt} {\kern 1pt} {\kern 1pt} {\kern 1pt} {\kern 1pt} {\kern 1pt}  - v_N^cx_N^w{\kern 1pt} {\kern 1pt} {\kern 1pt} {\kern 1pt} {\kern 1pt} {\kern 1pt} {\kern 1pt} {\kern 1pt} {\kern 1pt} {\kern 1pt} {\kern 1pt}  - v_N^cy_N^w{\kern 1pt} {\kern 1pt} {\kern 1pt} {\kern 1pt} {\kern 1pt} {\kern 1pt} {\kern 1pt} {\kern 1pt} {\kern 1pt}  - v_N^cz_N^w{\kern 1pt} {\kern 1pt} {\kern 1pt} {\kern 1pt} {\kern 1pt} {\kern 1pt} {\kern 1pt} {\kern 1pt} {\kern 1pt} {\kern 1pt} {\kern 1pt}  - v_N^c
\end{array} \right]\left[ \begin{array}{l}
a_{_{11}}^c\\
a_{_{12}}^c\\
a_{_{13}}^c\\
a_{_{14}}^c\\
a_{_{21}}^c\\
a_{_{22}}^c\\
a_{_{23}}^c\\
a_{_{24}}^c\\
a_{_{31}}^c\\
a_{_{32}}^c\\
a_{_{33}}^c\\
a_{_{34}}^c
\end{array} \right] = 0.
\end{aligned}
\end{equation}

The above formula can be written as  

\begin{equation} \label{eq17}
\begin{aligned}
AF = 1
\end{aligned},
\end{equation}

When N=6, the systems of linear equations can be solved directly.

When N$\le$6, we can obtain the least-squares solution under the constraint of $|F| = 1$. And then through SVD decomposition, the last column of the V matrix is the solution. 

\begin{equation} \label{eq18}
\begin{aligned}
F = UD{V^T}.
\end{aligned}
\end{equation}

By substituting Eq. 9 into Eq. 18, we can obtain Eq. 19.

\begin{equation} \label{eq19}
\begin{aligned}
F = \left[ {{I^c}{R^c}{\kern 1pt} {\kern 1pt} {\kern 1pt} {\kern 1pt} {\kern 1pt} {\kern 1pt} {I^c}{T^c}} \right].
\end{aligned}
\end{equation}

Thus, the rotation matrix $R^c$ and translation matrix $T^c$ can be calculated as follows.

\begin{equation} \label{eq20}
\begin{aligned}
{R^c} = {I^{ - 1}}\left[ \begin{array}{l}
a_{_{11}}^c{\kern 1pt} {\kern 1pt} {\kern 1pt} {\kern 1pt} a_{_{12}}^c{\kern 1pt} {\kern 1pt} {\kern 1pt} a_{_{13}}^c\\
a_{_{21}}^c{\kern 1pt} {\kern 1pt} {\kern 1pt} {\kern 1pt} a_{_{22}}^c{\kern 1pt} {\kern 1pt} {\kern 1pt} a_{_{23}}^c\\
a_{_{31}}^c{\kern 1pt} {\kern 1pt} {\kern 1pt} {\kern 1pt} a_{_{32}}^c{\kern 1pt} {\kern 1pt} {\kern 1pt} a_{_{33}}^c{\kern 1pt} 
\end{array} \right]
\end{aligned},
\end{equation}

\begin{equation} \label{eq21}
\begin{aligned}
{T^c} = {I^{ - 1}}\left[ \begin{array}{l}
{a_{14}}\\
{a_{24}}\\
{a_{34}}
\end{array} \right]
\end{aligned},
\end{equation}

After the rotation matrix and translation matrix are calculated, we completed the live camera localization for the current point data. 

\section{Experiments}
To investigate the performance of the proposed method, we conducted experiments on metrically consistent and indoor scenes. All experiments were implemented on a PC with AMD Ryzen 5 5600H (3GHz) CPU, and a packaged FPP sensor. The built FPP sensor is shown in Fig. 5, which contains a Basler acA1920-155uc camera (1920 $\times$ 1200 resolution), a LighterCraf ter 4500 (912 $\times$ 1140 resolution) and an interface expansion board. The interface expansion board is used to realize the synchronous triggering of the projector and the camera. The imaging speed of the sensor is 300 fps, the single-view measurement accuracy is 0.067 mm and the global mapping accuracy is around one millimeter. 

\begin{figure}[htbp] 
 \center{\includegraphics[width=9cm]  {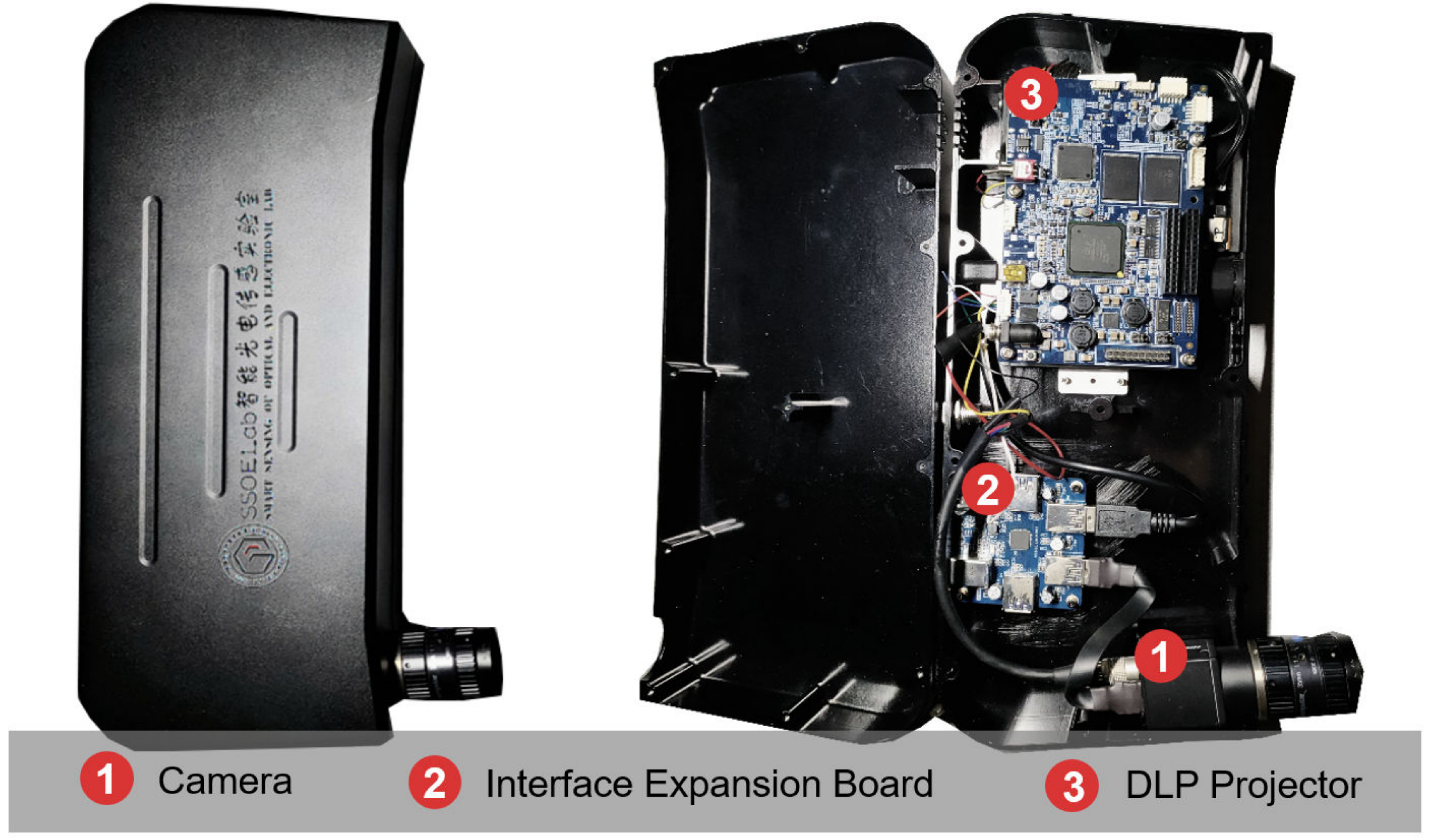}} 
 \caption{\label{1} Labeled diagram of FPP sensor.} 
\end{figure}

Firstly, a doll was measured to investigate the ability of metrically consistent models from trajectories containing loop closures. As shown in Fig. 6, the FPP system was placed in a fixed location and the doll was placed on a turntable. Then, the turntable was spun through a full rotation to capture point clouds from different perspectives of the doll for registration, and the trajectory localized by the corresponding camera should also be a circle and return to the origin, which allows us to easily evaluate the quality of camera localization. The main motivation for our experiments is to investigate the convergence of the mapping and localization schemes. As shown in the enlarged image of Fig. 6, the captured single-view point clouds were well registered, which demonstrates the performance of our mapping method. Then, the resulting sensor trajectories are given at the bottom of Fig. 6. Since the FPP sensor is fixed, the object is placed on the turntable, which can be equivalent to the sensor rotating around the object. In this experiment, the turntable was rotated 3 times, 13 times for each turn. The loop closures effect of localization trajectory performs well. The camera forms a perfectly circular trajectory around the doll and can return to the original location after one rotation, which demonstrates the performance of our localization method.

\begin{figure}[htbp] 
 \center{\includegraphics[width=12cm]  {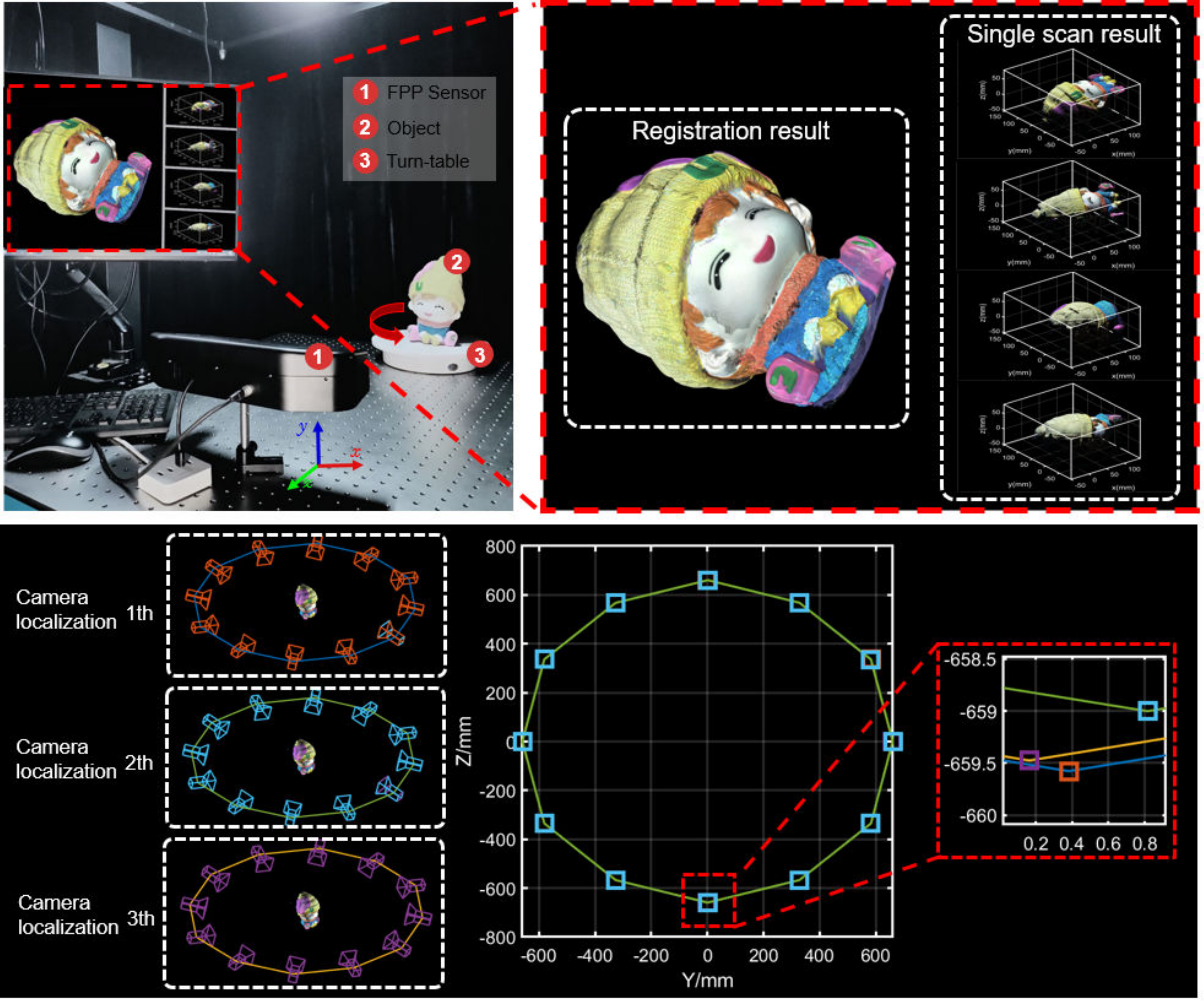}} 
 \caption{\label{1} Experimental results of the doll.} 
\end{figure}

Finally, to investigate the performance of our method for real scenes, we conducted the mapping and localization experiment in the corridor. The system setup of the indoor scene experiment is shown in Fig. 7, which includes a computer, our FPP sensor, a battery and a mobile cart. The FPP sensor is located in the mobile cart to scan the indoor scene. The results of scenes using our system are shown in Fig. 8. To generate this map of the indoor scene, we captured over 100 frames of single-view point clouds. The overlap area of adjacent point clouds accounts for one-third of the total area. As shown at the top of Fig. 8, the generated map point cloud is dense and has high-quality local geometric information. Then, the map with texture information is shown at the top of Fig. 8, which achieves perfect registration of multi-frame point clouds and avoids interference of cumulative errors. This also demonstrates that our global pose alignment strategy scales well to large spatial extents and long sequences.

\begin{figure}[htbp] 
 \center{\includegraphics[width=9cm]  {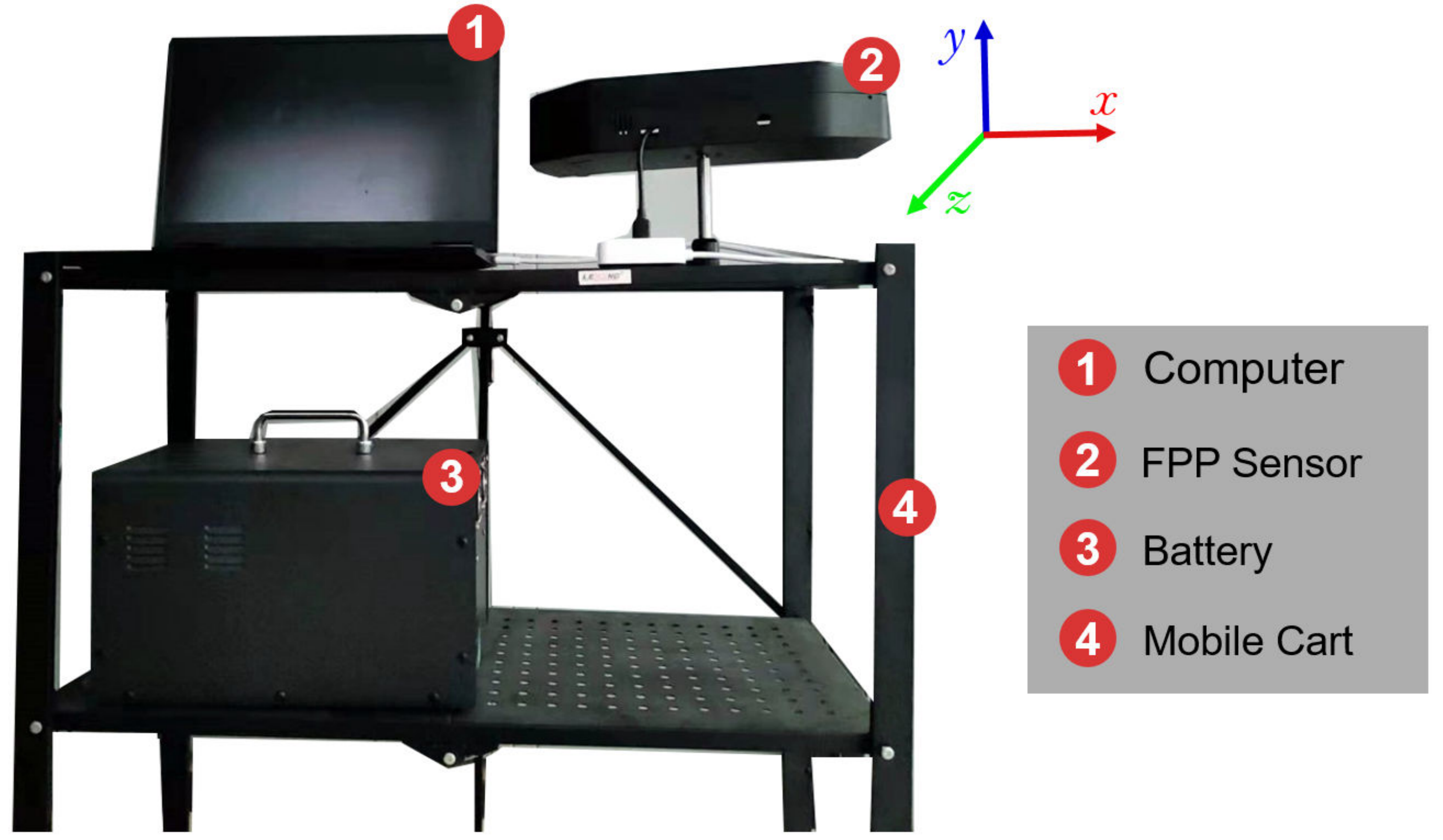}} 
 \caption{\label{1} The system setup of indoor scene experiment.} 
\end{figure}

\begin{figure}[htbp] 
 \center{\includegraphics[width=15.5cm]  {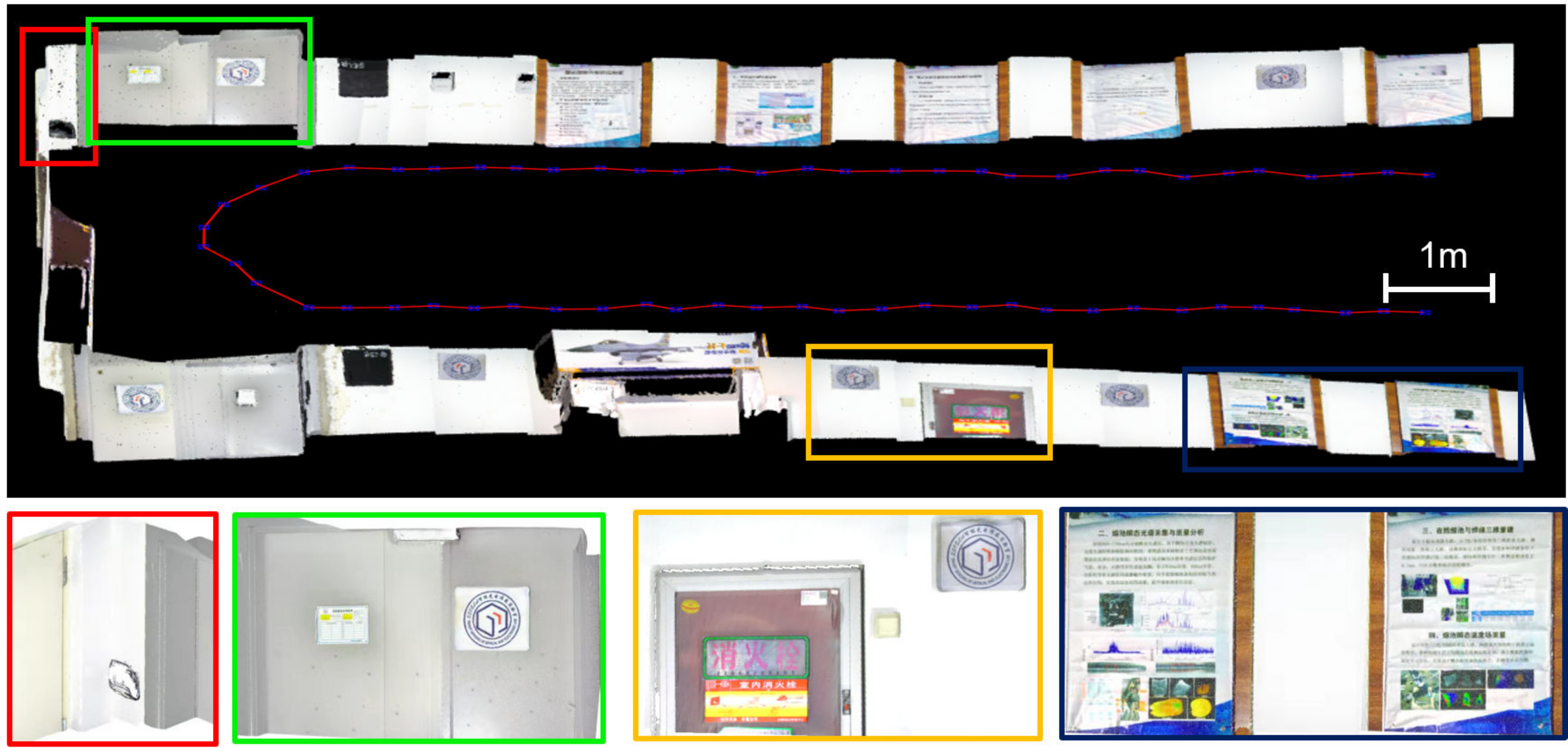}} 
 \caption{\label{1} Experimental results of the indoor scene.} 
\end{figure}

\section{Conclusion}
The availability of accurate 3-D metrology technologies such as FPP has the potential to revolutionise the fields of indoor localization and mapping, but methods to date have not fully leveraged the accuracy and speed of sensing that such techniques offer. This paper presents a millimeter (mm)-level indoor FPP-based SLAM method, which includes the front-end using FPP technique to perceive the unknown scene, and the correspond back-end algorithm using the 2D-to-3D descriptor and the coordinate transformation relationship of FPP to estimate of the pose of sensors and construct the map of indoor scene. The capabilities of the proposed method have been demonstrated on room sized SLAM. We believe that this paper will open a new door for the application of FPP in indoor SLAM.

\section*{Data Availability}
The data that support the findings of this study are available from the corresponding author upon reasonable request.

\section*{Declaration of competing interest}
The authors declare that they have no known competing financial interests or personal relationships that could have appeared to influence the work reported in this paper.

\bibliography{mybibfile}

\end{document}